\documentclass{article} 
\usepackage{nips15submit_e,times}
\usepackage{hyperref}
\usepackage{url}

\usepackage{graphicx}
\usepackage{amsmath}

\graphicspath{{figures/}}

\title{Spiking Deep Networks with LIF Neurons}

\author{
Eric Hunsberger \\
Centre for Theoretical Neuroscience \\
University of Waterloo \\
Waterloo, ON N2L 3G1 \\
\texttt{ehunsber@uwaterloo.ca} \\
\And
Chris Eliasmith \\
Centre for Theoretical Neuroscience \\
University of Waterloo \\
Waterloo, ON N2L 3G1 \\
\texttt{celiasmith@uwaterloo.ca} \\
}

\nipsfinalcopy 

\begin{document}

\maketitle

\begin{abstract}
We train spiking deep networks using leaky integrate-and-fire (LIF) neurons,
and achieve state-of-the-art results for spiking networks
on the CIFAR-10 and MNIST datasets.
This demonstrates that biologically-plausible spiking LIF neurons
can be integrated into deep networks
can perform as well as other spiking models (e.g. integrate-and-fire).
We achieved this result by softening the LIF response function,
such that its derivative remains bounded,
and by training the network with noise
to provide robustness against the variability introduced by spikes.
Our method is general and could be applied to other neuron types,
including those used on modern neuromorphic hardware.
Our work brings more biological realism into modern image classification models,
with the hope that these models
can inform how the brain performs this difficult task.
It also provides new methods for training deep networks
to run on neuromorphic hardware,
with the aim of fast, power-efficient image classification
for robotics applications.
\end{abstract}

\section{Introduction}

Deep artificial neural networks (ANNs) have recently been very successful
at solving image categorization problems.
Early successes with the MNIST database \cite{Lecun1998, Hinton2006}
were expanded to the more difficult but similarly sized
CIFAR-10 dataset \cite{Krizhevsky2010} and
Street-view house numbers dataset \cite{Sermanet2012}.
More recently, many groups have achieved better results
on these small datasets (e.g. \cite{Lee2015})
and as well as success on larger datasets (e.g. \cite{Gens2012}).
This work culminated with the application of deep neural networks
to ImageNet \cite{Krizhevsky2012},
a very large and challenging dataset.

The relative success of deep ANNs in general---%
and convolutional neural networks in particular---%
on these datasets have put them well ahead
of other methods in terms of image categorization by machines.
Given that deep ANNs are approaching human performance on some datasets
(or even passing it, for example on MNIST)
suggests that these models may be able to shed light
on how the human visual system solves these same tasks.

There has recently been considerable effort to take deep ANNs
and make them more biologically plausible by introducing neural ``spiking''
\cite{Brader2007, Eliasmith2012a, Neftci2013, O'Connor2013, Cao2014, Diehl2015},
such that connected nodes in the network transmit information
via instantaneous single bits (spikes),
rather than transmitting real-valued activities.
While one goal of this work is to better understand the brain
by trying to reverse engineer it \cite{Eliasmith2012a},
another goal is to build energy-efficient neuromorphic systems
that use a similar communication method for image categorization
\cite{Cao2014, Diehl2015}.

\section{Methods}

We first train a network on static images
using traditional deep learning techniques;
we call this the \emph{static network}.
We then take the parameters (weights and biases) from the static network
and use them to connect spiking neurons,
forming the \emph{dynamic network} (or spiking network).
The challenge is to train the static network in such a way that
a) it \emph{can} be transferred into a spiking network, and
b) the classification error of the dynamic network
is as close to that of the static network as possible
(this means the error rate is as low as possible,
since we do not expect the dynamic network
to perform better than the static one).

\subsection{Static convolutional network}

We base our network off that of Krizhevsky et al. \cite{Krizhevsky2012},
which achieved ~11\% error on the CIFAR-10 dataset
(a larger variant of the model won the ImageNet 2012 competition).
The original network consists of five layers:
two generalized convolutional layers,
followed by two locally-connected non-convolutional layers,
followed by a fully-connected softmax classifier.
A generalized convolutional layer consists of a set of convolutional weights
followed by a neural nonlinearity, then a pooling layer,
and finally a local response normalization layer.
The locally-connected non-convolutional layers
are also followed by a neural nonlinearity.
In the case of the original network,
the nonlinearity is a rectified linear (ReLU) function,
and both pooling layers perform overlapping max-pooling.
Code for the original network and details of the network architecture
and training
can be found at \url{https://code.google.com/p/cuda-convnet2/}.

To make the static network transferable to spiking neurons,
a number of modifications are necessary.
First, we remove the local response normalization layers.
This computation would likely require some sort of lateral connections
between neurons,
which are difficult to add in the current framework
since the resulting network would not be feedforward.

Second, we changed the pooling layers from max pooling to average pooling.
Again, computing max pooling would likely require lateral connections
between neurons,
making it difficult to implement without significant changes
to the training software.
While the Neural Engineering Framework
can be used to compute a max function in a feedforward manner \cite{Eliasmith2003},
this method requires prohibitively many neurons to achieve reasonable accuracy.
Average pooling, on the other hand, is very easy to compute in spiking neurons,
since it is simply a weighted sum.

The other modifications---%
using leaky integrate-and-fire neurons and training with noise---%
are the main focus of this paper,
and are described in detail below.

\subsection{Leaky integrate-and-fire neurons}

Our network uses a modified leaky integrate-and-fire (LIF) neuron nonlinearity
instead of the rectified linear nonlinearity.
Past work has kept the rectified linear nonlinearity for the static network
and substituted in the spiking integrate-and-fire (IF) neuron model
in the dynamic network \cite{Cao2014, Diehl2015},
since the static firing curve of the IF neuron model is a rectified line.
Our motivations for using the LIF neuron model are that
a) it is more biologically realistic than the IF neuron model \cite[p. 338]{Koch1999},
and b) it demonstrates that alternative models can be used in such networks.
The methods applied here are transferable to other neuron types,
and could be used to train a network for the idiosyncratic
neuron types employed by some neuromorphic hardware (e.g. \cite{Benjamin2014}).

The LIF neuron dynamics are given by the equation
\begin{align}
  \tau_{RC} \frac{dv(t)}{dt} = -v(t) + J(t)
  \label{eqn:lifode}
\end{align}
where $v(t)$ is the membrane voltage, $J(t)$ is the input current,
and $\tau_{RC}$ is the membrane time constant.
When the voltage reaches $V_{th} = 1$, the neuron fires a spike,
and the voltage is held at zero for a refractory period of $\tau_{ref}$.
Once the refractory period is finished,
the neuron obeys Equation~\ref{eqn:lifode} until another spike occurs.

Given a constant input current $J(t) = j$,
we can solve Equation~\ref{eqn:lifode} for the time it takes the voltage
to rise from zero to one,
and thereby find the steady-state firing rate
\begin{align}
  r(j) = \begin{cases}
    \left[\tau_{ref} - \tau_{RC} \log(1 - \frac{V_{th}}{j}) \right]^{-1} & \text{if } j > V_{th} \\
    0 & \text{otherwise}
  \end{cases}.
  \label{eqn:lifss}
\end{align}

Theoretically, we should be able to train a deep neural network
using Equation~\ref{eqn:lifss} as the static nonlinearity
and make a reasonable approximation of the network in spiking neurons,
assuming that the spiking network has a synaptic filter that sufficiently
smooths a spike train to give a good approximation of the firing rate.
The LIF steady state firing rate has the particular problem
that the derivative approaches infinity as $j \to 0_{+}$,
which causes problems when employing backpropagation.
To address this, we added smoothing to the LIF rate equation.

\begin{figure}
  \centering
  \includegraphics[width=\columnwidth]{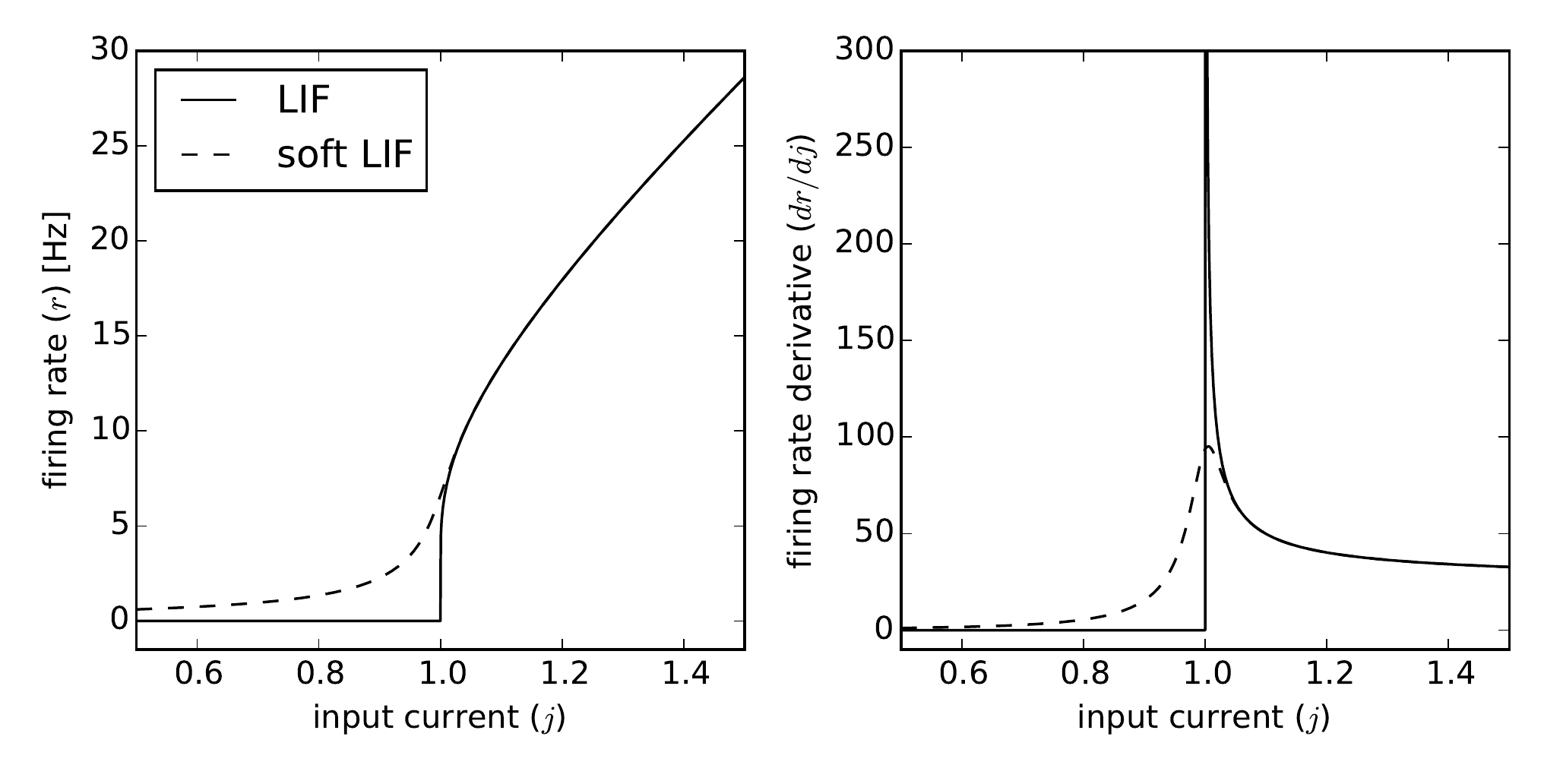}
  \caption{
    Comparison of LIF and soft LIF response functions.
    The left panel shows the response functions themselves.
    The LIF function has a hard threshold at $j = V_{th} = 1$;
    the soft LIF function smooths this threshold.
    The right panel shows the derivatives of the response functions.
    The hard LIF function has a discontinuous and unbounded derivative
    at $j = 1$; the soft LIF function has a continuous bounded derivative,
    making it amenable to use in backpropagation.
  }
  \label{fig:softlif}
\end{figure}

Equation~\ref{eqn:lifss} can be rewritten as
\begin{align}
  r(j) =
    \left[\tau_{ref} + \tau_{RC} \log(1 + \frac{V_{th}}{\rho(j - V_{th})}) \right]^{-1}
\end{align}
where $\rho(x) = \max(x, 0)$.
If we replace this hard maximum with a softer maximum $\rho_1(x) = \log(1 + e^x)$,
then the LIF neuron loses its hard threshold
and the derivative becomes bounded.
Further, we can use the substitution
\begin{align}
  \rho_2(x) = \gamma \log\left[1 + e^{x / \gamma}\right]
  \label{eqn:softrelusigma}
\end{align}
to allow us control over the amount of smoothing,
where $\rho_2(x) \to \max(x, 0)$ as $\gamma \to 0$.
Figure~\ref{fig:softlif} shows the result of this substitution.

\subsection{Training with noise}

Training neural networks with various types of noise on the inputs
is not a new idea.
Denoising autoencoders \cite{Vincent2008} have been successfully applied
to datasets like MNIST,
learning more robust solutions with lower generalization error than
their non-noisy counterparts.

In a spiking neural network, the neuron receiving spikes in a connection
(called the post-synaptic neuron)
actually receives a filtered version of each spike.
This filtered spike is called a post-synaptic current (or potential),
and the shape of this signal is determined
by the combined dynamics of the pre-synaptic neuron
(e.g. how much neurotransmitter is released)
and the post-synaptic neuron
(e.g. how many ion channels are activated by the neurotransmitter
and how they affect the current going into the neuron).
This post-synaptic current dynamics can be characterized relatively well
as a linear system with the impulse response given by the $\alpha$-function
\cite{Mainen1995}:
\begin{align}
  \alpha(t) = \frac{t}{\tau_s} e^{-t / \tau_s}.
  \label{eqn:alpha}
\end{align}

The filtered spike train can be viewed as an estimate of the neuron activity.
For example, if the neuron is firing regularly at 200 Hz,
filtering spike train will result in a signal fluctuating around 200 Hz.
We can view the neuron output as being 200 Hz,
with some additional ``noise'' around this value.
By training our static network with some random noise
added to the output of each neuron for each training example,
we can simulate the effects of using spikes
on the signal received by the post-synaptic neuron.

\begin{figure}
  \centering
  \includegraphics[width=0.8\columnwidth]{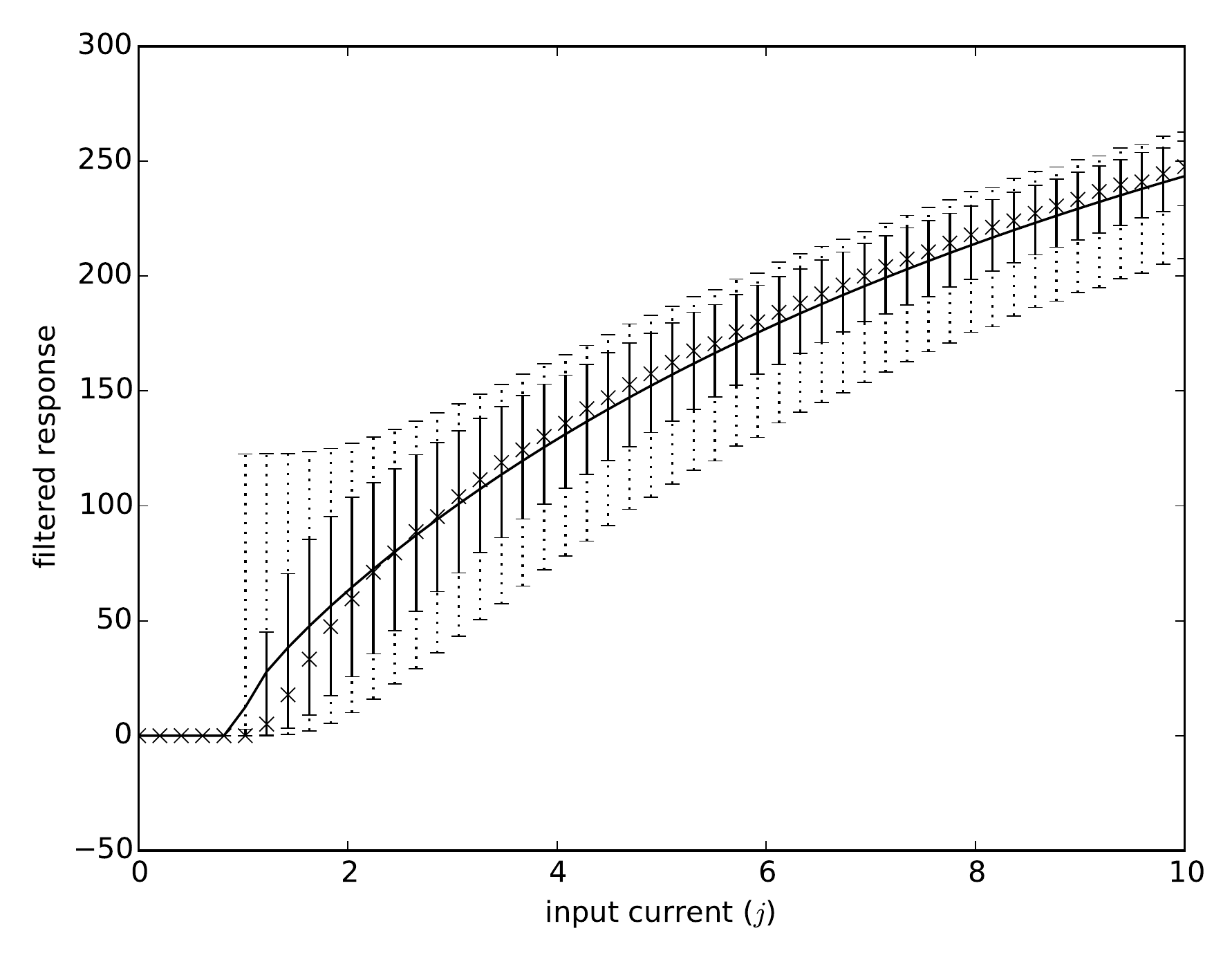}
  \caption{Variability in filtered spike trains versus
    input current for the LIF neuron
    ($\tau_{RC} = 0.02, \tau_{ref} = 0.004$).
    The solid line shows the mean of the filtered spike train
    (which matches the analytical rate of Equation~\ref{eqn:lifss}),
    the `x'-points show the median, the solid error bars show the 25th and 75th
    percentiles,
    and the dotted error bars show the minimum and maximum.
    The spike train was filtered with an $\alpha$-filter
    (Equation~\ref{eqn:alpha}) with $\tau_s = 0.003$ s.
    (Note that this is different than the $\tau_s = 0.005$ used in simulation,
    to better display the variation.)
  }
  \label{fig:noise}
\end{figure}

Figure~\ref{fig:noise} shows how the variability of filtered spike trains
depends on input current for the LIF neuron.
Since the impulse response of the $\alpha$-filter has an integral of one,
the mean of the filtered spike trains is equal to the analytical rate
of Equation~\ref{eqn:lifss}.
However, the statistics of the filtered signal vary significantly
across the range of input currents.
Just above the firing threshold,
the distribution is skewed towards higher firing rates
(i.e. the median is below the mean),
since spikes are infrequent so the filtered signal has time to return
to near zero between spikes.
At higher input currents, on the other hand,
the distribution is skewed towards lower firing rates
(i.e. the median is above the mean).
In spite of this,
we used a Gaussian distribution to generate the additive noise during training,
for simplicity.
We found the average standard deviation to be approximately $\sigma = 10$
across all positive input currents for an $\alpha$-filter with $\tau_s = 0.005$.
The final steady-state soft LIF curve used in training is given by
\begin{align}
  r(j) =
    \left[\tau_{ref} +
      \tau_{RC} \log(1 + \frac{V_{th}}{\rho(j - V_{th})}) \right]^{-1}
    + \eta(j)
  \label{eqn:softlifnoise}
\end{align}
where
\begin{align}
  \eta(j) \sim \begin{cases}
    G(0, \sigma) & \text{if } j > V_{th} \\
    0 & \text{otherwise}
  \end{cases}
\end{align}
and $\rho(\cdot)$ is given by Equation~\ref{eqn:softrelusigma}.

\subsection{Conversion to a spiking network}

Finally, we convert the trained static network
to a dynamic spiking network.
The parameters in the spiking network (i.e. weights and biases)
are all identical to that of the static network.
The convolution operation also remains the same,
since convolution can be rewritten as simple connection weights (synapses)
$w_{ij}$ between pre-synaptic neuron $i$ and post-synaptic neuron $j$.
(How the brain might \emph{learn} connection weight patterns, i.e. filters,
that are repeated at various points in space,
is a much more difficult problem that we will not address here.)
Similarly, the average pooling operation can be written
as a simple connection weight matrix,
and this matrix can be multiplied by the convolutional weight matrix
of the following layer to get direct connection weights between
neurons.\footnote{For computational efficiency,
we actually compute the convolution and pooling separately.}

The only component of the network that actually changes, then,
when moving from the static to the dynamic network,
is the neurons themselves.
The most significant change is that we replace
the soft LIF rate model (Equation~\ref{eqn:softlifnoise})
with the LIF spiking model (Equation~\ref{eqn:lifode}).
We also remove the additive Gaussian noise used in training.

Additionally, we add post-synaptic filters to the neurons,
which filter the incoming spikes before passing the resulting currents
to the LIF neuron equation.
As stated previously, we use the $\alpha$-filter for our synapse model,
since it has both strong biological support \cite{Mainen1995},
and removes a significant portion of the high-frequency variation
produced by spikes.
We pick the decay time constant $\tau_s = 5$ ms,
typical for excitatory AMPA receptors in the brain \cite{Jonas1993}.

\section{Results}

We tested our network on the CIFAR-10 dataset.
This dataset is composed of 60000 $32 \times 32$ pixel labelled images
from ten categories.
We used the first 50000 images for training and the last 10000 for testing,
and augmented the dataset by taking random $24 \times 24$ patches from the
training images
and then testing on the center patches from the testing images.
This methodology is similar to Krizhevsky et al. \cite{Krizhevsky2012},
except that they also used multiview testing where the classifier output
is the average output of the classifier run on nine random patches from
each testing image (increasing the accuracy by about 2\%).

Table~\ref{tab:mods} shows the effect of each modification
on the network classification error.
Our original static network based on the methods of \cite{Krizhevsky2012}
achieved 14.63\% error,
which is higher than the ~11\% achieved by the original paper since
a) we are not using multiview testing,
and b) we used a shorter training time (160 epochs versus 520 epochs).

\begin{table}
  \centering
  \begin{minipage}{11.5cm}
    \begin{center}
    \begin{tabular}{|c|l|c|}\hline
      \# & Modification & CIFAR-10 error \\\hline\hline
      0 & Original static network based on Krizhevsky et al. \cite{Krizhevsky2012} & 14.63\% \\\hline
      1 & Above minus local contrast normalization & 15.27\% \\\hline
      2 & Above minus max pooling & 17.20\% \\\hline
      3 & Above with soft LIF & 18.92\% \\\hline
      4 & Above with training noise ($\sigma = 10$) & 19.74\% \\\hline
      5 & Above with training noise ($\sigma = 20$) & 22.22\% \\\hline
      6 & Network 3 ($\sigma = 0$) in spiking neurons & 25.1\%$^a$ \\\hline
      7 & Network 4 ($\sigma = 10$) in spiking neurons & 21.7\%$^a$ \\\hline
      8 & Network 5 ($\sigma = 20$) in spiking neurons & 23.2\%$^a$ \\\hline
      9 & Network 4 ($\sigma = 10$) with additional training epochs & 16.01\% \\\hline
      10 & Network 9 ($\sigma = 10$) in spiking neurons & \textbf{17.05\%} \\\hline
    \end{tabular}
    \end{center}
    $^a$ Results from the same random 1000-image subset of the testing set.
  \end{minipage}
  \caption{Effects of successive modifications to CIFAR-10 error.
    We first show the original static (non-spiking) network based on \cite{Krizhevsky2012}.
    Modifications 1-5 are cumulative, which each one applied in addition to the previous ones.
    Rows 6-8 show the results of running static networks 3-5
    in spiking neurons, respectively.
    Row 9 shows the best architecture for spiking implementation,
    Network 4, trained for additional epochs,
    and row 10 shows this highly-trained network in spiking neurons.
    This is the best spiking-network result on CIFAR-10 to date.
  }
  \label{tab:mods}
\end{table}

Rows 1-5 in Table~\ref{tab:mods} show that
each successive modification to make the network amenable to running
in spiking neurons adds about 1-2\% more error.
Despite the fact that training with noise adds additional error
to the static network,
rows 6-8 of the table show that in the spiking network,
training with noise pays off,
though training with too much noise is not advantageous.
Specifically, though training with $\sigma = 20$ versus $\sigma = 10$
decreased the error introduced when switching to spiking neurons
(~1\% versus ~2\%),
training with $\sigma = 20$ versus $\sigma = 10$ introduced
an additional ~2.5\% error to the static network,
making the final spiking network perform worse.
In the interest of time, these spiking networks were all run
on the same 1000-image random subset of the testing data.
The last two rows of the table show the network
with the optimal amount of noise ($\sigma = 10$)
trained for additional epochs (a total of 520 as opposed to 160),
and run on the entire test set.
Our spiking network achieves an error of 17.05\% on the full CIFAR-10 test set,
which is the best published result of a spiking network on this dataset.

Comparing spiking networks is difficult,
since the results depend highly on the characteristics of the neurons used.
For example, neurons with very high firing rates, when filtered,
will result in spiking networks that behave almost identically
to their static counterparts.
Using neurons with lower firing rates
have much more variability in their filtered spike trains,
resulting in noisier and less accurate dynamic networks.
Nevertheless, we find it worthwhile to compare our results with those of
Cao et al. \cite{Cao2014},
who achieved 22.57\% error on the CIFAR-10 dataset
(as far as we know, the only other spiking network with published results on
CIFAR-10).
Our approach is in many ways similar to theirs,
with the notable differences that we used the LIF neuron
instead of the IF neuron,
and that we used noise during training.
The fact that we achieved marginally better results suggests that
LIF neuron spiking networks can be trained to state-of-the-art accuracy
and that adding noise during training helps improve accuracy.

We also measured the spike rates for our network on the CIFAR-10 dataset.
The average firing rate across all neurons in the network was
148 spikes/s, estimated from 20 test examples (30 seconds of simulated time).
The firing rate was quite different between different layers of the network,
with the first two convolutional layers having average firing rates of
172 spikes/s and 104 spikes/s respectively,
and the locally connected layers having rates of
10.6 spikes/s and 7.6 spikes/s respectively.
Cao et al. \cite{Cao2014} reported the number of
post-synaptic spikes ($5 \times 10^5$) for their network on the Tower dataset.
We used this, along with their
simulation time (100 ms) and number of neurons (57606),
to estimate the average firing rate of their network at 86.8 spikes/s.
However, their firing rates on the CIFAR-10 dataset could be significantly different,
since firing rates are heavily dependent on learned model parameters
(i.e. weights and biases) which can vary significantly between datasets.

Most spiking deep networks to date have been tested on the MNIST dataset.
The MNIST dataset is composed of 70000 labelled hand-written digits,
with 60000 used for training and 10000 reserved for testing.
While this dataset is quickly becoming obsolete as deep networks become
more and more powerful,
it is only recently that spiking networks are beginning to achieve
human-level accuracy on the dataset.

We trained an earlier version of our network on the MNIST dataset.
This version used layer-wise pretraining of non-convolutional denoising autoencoders,
stacked and trained as a classifier.
This network had two hidden layers of 500 and 200 nodes each,
and was trained on the unaugmented dataset.
Despite the significant differences between this network
and the network used on the CIFAR-10 dataset,
both networks use spiking LIF neurons and are trained with noise
to minimize the error caused by the filtered spike train variation.
Table~\ref{tab:mnist} shows a comparison between our network
and the best published results on MNIST.
Our network significantly outperforms the best results using LIF neurons,
and is on par with those of IF neurons.
This demonstrates that state-of-the-art networks can be trained with LIF neurons.
The average firing rate of this network is 25.7 spikes/s,
with the hidden layers averaging 23.0 spikes/s and 32.5 spikes/s, respectively.

\begin{table}
  \centering
  \begin{minipage}{11.5cm}
    \renewcommand*\footnoterule{\vspace{-0.5em}}
    \centering
    \begin{tabular}{|l|l|}
      \hline
      Source & MNIST error \\\hline\hline
      Brader et al. \cite{Brader2007} & 3.5\%
      (1.3\% misclassified, 2.2\% not classified) (IF) \\\hline
      Eliasmith et al. \cite{Eliasmith2012a} & 6\% (LIF) \\\hline
      Neftci et al. \cite{Neftci2013} & 8.1\% (LIF) \\\hline
      O'Connor et al. \cite{O'Connor2013} & 2.52\% (sigmoid-binary), 5.91\% (LIF)\\\hline
      Garbin et al. \cite{Garbin2014} & 1.7\% (IF) \\\hline
      Diehl et al. \cite{Diehl2015} & 1.36\%
      \footnote{Their best result for a non-convolutional network.} (IF) \\\hline
      Our network & 1.63\% (LIF) \\\hline
    \end{tabular}
  \end{minipage}
  \caption{
    Comparison of our network to the best published results
    for spiking networks on MNIST.
    Our network performs on par with state-of-the-art results,
    demonstrating that state-of-the-art spiking networks can be trained with LIF neurons.
  }
  \label{tab:mnist}
\end{table}

\section{Discussion}

Our results demonstrate that it is possible to train
accurate deep convolutional networks for image classification
using more biologically accurate leaky integrate-and-fire (LIF) neurons,
as opposed to the traditional rectified-linear or sigmoid neurons.
Such a network can be run in spiking neurons,
and training with noise decreases the amount of error introduced
when running in spiking versus rate neurons.

The first main contribution of this paper is to demonstrate that
state-of-the-art deep spiking networks can be trained with LIF neurons.
Other state-of-the-art methods use integrate-and-fire (IF) neurons
\cite{Cao2014, Diehl2015},
which are easier to fit to the rectified linear units commonly used
in deep convolutional networks,
but are biologically implausible.
By smoothing the LIF response function so that its derivative remains bounded,
we are able to use this more biologically plausible neuron
with a standard convolutional network trained by backpropagation.

This idea of smoothing the neuron response function is applicable
to other neuron types, too.
Many other neuron types have discontinuous response functions
(e.g. the FitzHugh-Nagumo neuron),
and our smoothing method allows such neurons to be used
in deep convolutional networks.
We found that there was very little error introduced
by switching from the soft response function to the hard response function
with LIF neurons for the amount of smoothing that we used.
However, for neurons with harsh discontinuities that require more smoothing,
it may be necessary to slowly relax the smoothing
over the course of the training
so that, by the end of the training, the smooth response function
is arbitrarily close to the hard response function.

The other main contribution of this paper is to demonstrate
that training with noise on neuron outputs
can decrease the error introduced when transitioning to spiking neurons.
Training with noise on neuron outputs improved the performance
of the spiking network considerably (the error decreased by 3.4\%).
This is because noise on the output of the neuron simulates the variability
that a spiking network encounters when filtering a spike train.
There is a tradeoff between too little training noise,
where the resultant dynamic network is not robust enough against spiking variability,
and too much noise,
where the accuracy of the static network is decreased.
Since the variability produced by spiking neurons is not Gaussian
(Figure~\ref{fig:noise}),
our additive Gaussian noise is a rough approximation of the variability
that the spiking network will encounter.
Future work includes training with noise that is more representative
of the variability seen in spiking networks,
to accommodate both the non-Gaussian statistics at any particular input current,
and the changing statistics across input currents.

Direct comparison with other spiking neural networks is difficult,
since the amount of error introduced when converting
from a static to a spiking network
is heavily dependent on the firing rates of the neurons.
Nevertheless, we found our network to perform favourably
with other spiking networks,
achieving the best published result for a spiking network on CIFAR-10,
and the best result for a LIF neuron spiking network on MNIST.
We also report our average firing rates for each layer
and for the entire network,
to facilitate comparison with future networks.
The firing rates for the convolutional layers of our network
are higher than typical in visual cortex \cite{Carandini2000}.
Future work includes looking at methods to lower firing rates,
though this may involve sparsification of neural firing---%
having fewer neurons fire for a particular stimulus---%
which can be difficult in convolutional networks.

Other future work includes implementing max-pooling
and local contrast normalization layers in spiking networks.
Networks could also be trained offline as described here
and then fine-tuned online using an STDP rule,
such as the one described in \cite{Nessler2013},
to help further reduce errors associated with converting from
rate-based to spike-based networks,
while avoiding difficulties with training
a network in spiking neurons from scratch.

\renewcommand{\refname}{\subsubsection*{References}}

{\small

}


\begin{thebibliography}{10}
\providecommand{\url}[1]{#1}
\csname url@samestyle\endcsname
\providecommand{\newblock}{\relax}
\providecommand{\bibinfo}[2]{#2}
\providecommand{\BIBentrySTDinterwordspacing}{\spaceskip=0pt\relax}
\providecommand{\BIBentryALTinterwordstretchfactor}{4}
\providecommand{\BIBentryALTinterwordspacing}{\spaceskip=\fontdimen2\font plus
\BIBentryALTinterwordstretchfactor\fontdimen3\font minus
  \fontdimen4\font\relax}
\providecommand{\BIBforeignlanguage}[2]{{%
\expandafter\ifx\csname l@#1\endcsname\relax
\typeout{** WARNING: IEEEtran.bst: No hyphenation pattern has been}%
\typeout{** loaded for the language `#1'. Using the pattern for}%
\typeout{** the default language instead.}%
\else
\language=\csname l@#1\endcsname
\fi
#2}}
\providecommand{\BIBdecl}{\relax}
\BIBdecl

\bibitem{Lecun1998}
Y.~Lecun, L.~Bottou, Y.~Bengio, and P.~Haffner, ``{Gradient-based learning
  applied to document recognition},'' \emph{Proceedings of the IEEE}, vol.~86,
  no.~11, pp. 2278--2324, 1998.

\bibitem{Hinton2006}
G.~E. Hinton and R.~R. Salakhutdinov, ``{Reducing the dimensionality of data
  with neural networks.}'' \emph{Science (New York, N.Y.)}, vol. 313, no. 5786,
  pp. 504--7, jul 2006.

\bibitem{Krizhevsky2010}
A.~Krizhevsky, ``{Convolutional deep belief networks on CIFAR-10},'' Tech.
  Rep., 2010.

\bibitem{Sermanet2012}
P.~Sermanet, S.~Chintala, and Y.~LeCun, ``{Convolutional neural networks
  applied to house numbers digit classification},'' in \emph{International
  Conference on Pattern Recognition (ICPR)}, 2012.

\bibitem{Lee2015}
C.-Y. Lee, S.~Xie, P.~W. Gallagher, Z.~Zhang, and Z.~Tu, ``{Deeply-supervised
  nets},'' in \emph{International Conference on Artificial Intelligence and
  Statistics (AISTATS)}, vol.~38, 2015, pp. 562--570.

\bibitem{Gens2012}
R.~Gens and P.~Domingos, ``{Discriminative learning of sum-product networks},''
  in \emph{Advances in Neural Information Processing Systems (NIPS)}, 2012, pp.
  1--9.

\bibitem{Krizhevsky2012}
A.~Krizhevsky, I.~Sutskever, and G.~Hinton, ``{Imagenet classification with
  deep convolutional neural networks},'' in \emph{Advances in Neural
  Information Processing Systems}, 2012.

\bibitem{Brader2007}
J.~M. Brader, W.~Senn, and S.~Fusi, ``{Learning real-world stimuli in a neural
  network with spike-driven synaptic dynamics},'' \emph{Neural Computation},
  vol.~19, pp. 2881--2912, 2007.

\bibitem{Eliasmith2012a}
C.~Eliasmith, T.~C. Stewart, X.~Choo, T.~Bekolay, T.~DeWolf, C.~Tang, and
  D.~Rasmussen, ``{A Large-Scale Model of the Functioning Brain},''
  \emph{Science}, vol. 338, no. 6111, pp. 1202--1205, nov 2012.

\bibitem{Neftci2013}
E.~Neftci, S.~Das, B.~Pedroni, K.~Kreutz-Delgado, and G.~Cauwenberghs,
  ``{Event-driven contrastive divergence for spiking neuromorphic systems},''
  \emph{Frontiers in Neuroscience}, vol.~7, no. 272, 2013.

\bibitem{O'Connor2013}
P.~O'Connor, D.~Neil, S.-C. Liu, T.~Delbruck, and M.~Pfeiffer, ``{Real-time
  classification and sensor fusion with a spiking deep belief network},''
  \emph{Frontiers in Neuroscience}, vol.~7, jan 2013.

\bibitem{Cao2014}
Y.~Cao, Y.~Chen, and D.~Khosla, ``{Spiking Deep Convolutional Neural Networks
  for Energy-Efficient Object Recognition},'' \emph{International Journal of
  Computer Vision}, vol. 113, no.~1, pp. 54--66, nov 2014.

\bibitem{Diehl2015}
P.~U. Diehl, D.~Neil, J.~Binas, M.~Cook, S.-C. Liu, and M.~Pfeiffer,
  ``{Fast-Classifying, High-Accuracy Spiking Deep Networks Through Weight and
  Threshold Balancing},'' in \emph{IEEE International Joint Conference on
  Neural Networks (IJCNN)}, 2015.

\bibitem{Eliasmith2003}
C.~Eliasmith and C.~H. Anderson, \emph{{Neural Engineering: Computation,
  Representation, and Dynamics in Neurobiological Systems}}.\hskip 1em plus
  0.5em minus 0.4em\relax Cambridge, MA: MIT Press, 2003.

\bibitem{Koch1999}
C.~Koch, \emph{{Biophysics of computation: Information processing in single
  neurons}}.\hskip 1em plus 0.5em minus 0.4em\relax New York, NY: Oxford
  University Press, 1999.

\bibitem{Benjamin2014}
B.~V. Benjamin, P.~Gao, E.~McQuinn, S.~Choudhary, A.~R. Chandrasekaran, J.-M.
  Bussat, R.~Alvarez-Icaza, J.~V. Arthur, P.~A. Merolla, and K.~Boahen,
  ``{Neurogrid: A mixed-analog-digital multichip system for large-scale neural
  simulations},'' \emph{Proceedings of the IEEE}, vol. 102, no.~5, pp.
  699--716, 2014.

\bibitem{Vincent2008}
P.~Vincent, H.~Larochelle, Y.~Bengio, and P.-A. Manzagol, ``{Extracting and
  composing robust features with denoising autoencoders},'' in
  \emph{International Conference on Machine Learning (ICML)}, 2008, pp.
  1096--1103.

\bibitem{Mainen1995}
Z.~F. Mainen and T.~J. Sejnowski, ``{Reliability of spike timing in neocortical
  neurons.}'' \emph{Science (New York, N.Y.)}, vol. 268, no. 5216, pp. 1503--6,
  jun 1995.

\bibitem{Jonas1993}
P.~Jonas, G.~Major, and B.~Sakmann, ``{Quantal components of unitary EPSCs at
  the mossy fibre synapse on CA3 pyramidal cells of rat hippocampus.}''
  \emph{The Journal of Physiology}, vol. 472, pp. 615--663, 1993.

\bibitem{Garbin2014}
D.~Garbin, O.~Bichler, E.~Vianello, Q.~Rafhay, C.~Gamrat, L.~Perniola,
  G.~Ghibaudo, and B.~DeSalvo, ``{Variability-tolerant convolutional neural
  network for pattern recognition applications based on OxRAM synapses},'' in
  \emph{IEEE International Electron Devices Meeting (IEDM)}, 2014, pp.
  28.4.1--28.4.4.

\bibitem{Carandini2000}
M.~Carandini and D.~Ferster, ``{Membrane potential and firing rate in cat
  primary visual cortex},'' \emph{The Journal of Neuroscience}, vol.~20, no.~1,
  pp. 470--484, 2000.

\bibitem{Nessler2013}
B.~Nessler, M.~Pfeiffer, L.~Buesing, and W.~Maass, ``{Bayesian computation
  emerges in generic cortical microcircuits through spike-timing-dependent
  plasticity.}'' \emph{PLoS computational biology}, vol.~9, no.~4, p. e1003037,
  apr 2013.

\end{thebibliography}
\end{document}